\documentclass[runningheads]{llncs}

\usepackage{eccv}

\usepackage{eccvabbrv}

\usepackage{graphicx}
\usepackage{booktabs}

\usepackage[accsupp]{axessibility}  

\usepackage[hidelinks]{hyperref}

\usepackage{orcidlink}

\usepackage{xcolor}
\usepackage{adjustbox}
\usepackage{multirow}
\usepackage{wrapfig}
\usepackage{bm}

\usepackage{pifont}

\definecolor{darkred}{RGB}{204, 0, 0}
\definecolor{darkgreen}{RGB}{0, 160, 0}

\DeclareMathOperator*{\argmax}{arg\,max}

\begin{document}

\title{Relightable Neural Actor with Intrinsic Decomposition and Pose Control} 

\titlerunning{Relightable Neural Actor with Intrinsic Decomposition and Pose Control}

\author{Diogo Carbonera Luvizon\inst{1,2} \and
Vladislav Golyanik\inst{1} \and
Adam Kortylewski\inst{1,3} \and\\
Marc Habermann\inst{1,2} \and
Christian Theobalt\inst{1,2}
}

\authorrunning{D.C.~Luvizon et al.}

\institute{
Max Planck Institute for Informatics, Saarland Informatics Campus \and
Saarbrücken Research Center for Visual Computing, Interaction and AI \and
University of Freiburg\\
Project page: \url{https://vcai.mpi-inf.mpg.de/projects/RNA}}

\maketitle

\begin{abstract}
Creating a controllable and relightable digital avatar from multi-view video with fixed illumination is a very challenging problem since humans are highly articulated, creating pose-dependent appearance effects, and skin as well as clothing require space-varying BRDF modeling.
Existing works on creating animatible avatars either do not focus on relighting at all, require controlled illumination setups, or try to recover a relightable avatar from very low cost setups, i.e. a single RGB video, at the cost of severely limited result quality, e.g. shadows not even being modeled. 
To address this, we propose \textit{Relightable Neural Actor}, a new video-based method for learning a pose-driven neural human model that can be relighted, allows appearance editing, and models pose-dependent effects such as wrinkles and self-shadows.
Importantly, for training, our method solely requires a multi-view recording of the human under a known, but static lighting condition.
To tackle this challenging problem, we leverage an implicit geometry representation of the actor with a drivable density field that models pose-dependent deformations and derive a dynamic mapping between 3D and UV spaces, where normal, visibility, and materials are effectively encoded.
To evaluate our approach in real-world scenarios, we collect a new dataset with four identities recorded under different light conditions, indoors and outdoors, providing the first benchmark of its kind for human relighting, and demonstrating state-of-the-art relighting results for novel human poses. 
\end{abstract}
%
%
\section{Introduction} \label{sec:intro}
Creating a relightable and controllable virtual actor of a real person solely from multi-view videos under static illumination is an exciting and challenging topic that has numerous applications in virtual and augmented reality, gaming, human-machine interaction, and content creation.
However, it requires modeling the outgoing radiance of a particular surface point on the human, which depends on the pose-dependent and highly non-rigid surface deformation of the human body, the complex and spatially varying materials, as well as the scene lighting. 
Intrinsically decomposing these quantities solely from multi-view video and enabling novel compositions that are of highly photorealistic quality defines an unsolved and challenging inverse problem in the literature.
\begin{figure}[tb]
    \centering
    \captionsetup{type=figure}
    \includegraphics[width=1.0\textwidth]{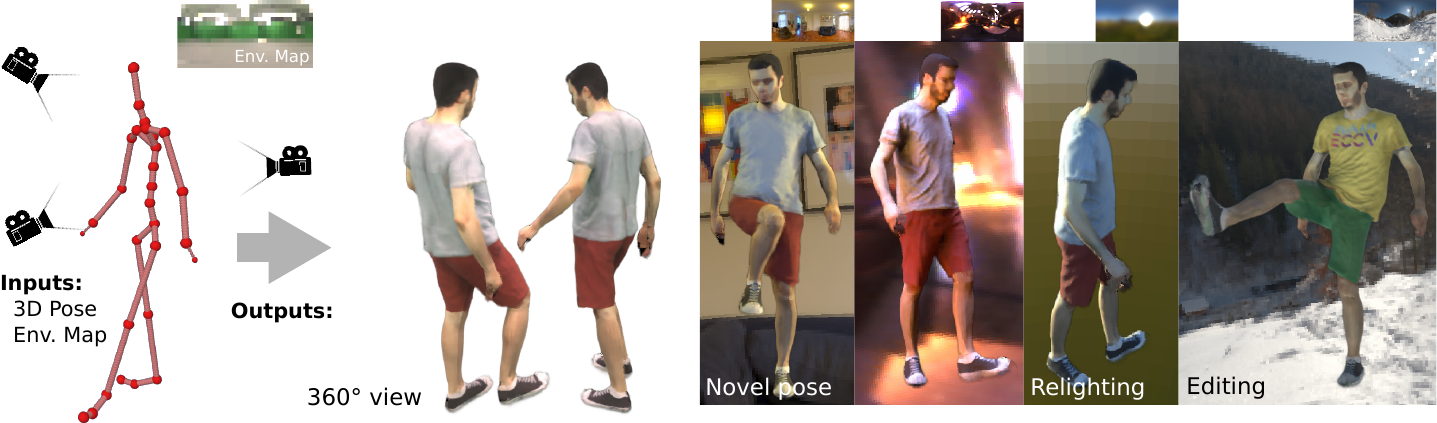}
    \captionof{figure}{
    \textbf{Our method learns a neural actor that is driven only by a 3D skeletal pose and allows rendering and editing of the actor's appearance under new lightning and poses not seen during training.} 
    Our approach models pose-dependent deformations, self-shadows, and performs intrinsic decomposition of dynamic humans through a pose-dependent mapping between 3D and UV spaces, which also enables editing the appearance and material properties at inference time.
    For training, our method only needs a multi-view video and an environment map.
    }
    \label{fig:teaser} 
\end{figure}
%
%
\par
Relighting \textit{static} scenes from multi-view imagery is a well-studied problem~\cite{zhang2021nerfactor,Srinivasan2020nerv,jin2023tensoir,liu2023nero}.
However, directly applying such methods to moving humans, which are dynamic, highly articulated, and have complex material composition, is not trivially possible, as these approaches do not account for dynamics. 
On the contrary, some works have focused on building drivable and dynamic full-body avatars from multi-view video~\cite{liu2021neural,peng2021_neural_body,ARAH_ECCV_2022} without considering relightability at all.
Despite some works~\cite{iwase2023relightablehands, guo2019relightables, DeepRelightable2020, bi2021deep, zhou2023relightable} assuming a \textit{well-calibrated lightstage settings}, only few human-specific works~\cite{chen2022_Relighting4D, zheng2023learning, iqbal2023rana} try to achieve human intrinsic decomposition and relighting from \textit{videos with static lights}.
While they are able to recover a relightable human representation from a single RGB video, they can solely replay the human motion under novel lighting conditions~\cite{chen2022_Relighting4D}, or do not model self-shadows and space-varying BRDF~\cite{iqbal2023rana}, which results in low visual fidelity.
In stark contrast, we assume a multi-view setting while accounting for high-frequency surface deformations as well as induced high-frequency lighting effects such as self-shadows and shading, which are indispensable for a truly immersive experience.
%
%
\par
To address this, we propose \textit{Relightable Neural Actor}, a new method that achieves intrinsic decomposition of dynamic human avatars, enabling pose-control, material editing, and realistic rendering with pose-dependent effects, while solely requiring a multi-view video of the real human under a single and static light condition during training (see \cref{fig:teaser}). At inference time, we only need an arbitrary driving skeletal pose and an environment map.
\par
We solve this challenging problem by leveraging an implicit density field from a pose-driven neural actor and by proposing a novel intrinsic decomposition guided by the implicit density field.
This strategy allows our method to model pose-dependent deformations in the geometry as dynamic maps, while representing the albedo and roughness properties of the actor as static maps.
Our approach not only makes the problem computationally tractable for the long and dynamic human sequences, but also enables editing the neural actor's appearance and materials at inference time.
Our key contribution is to sample normal and visibility from the posed density field and aggregate them into sparse UV maps, which are then densified by the proposed NormalNet and VisibilityNet models. 
This allows our approach to effectively model wrinkles and self-shadows even for long sequences.
The albedo and roughness maps are optimized during training and sampled by the proposed UVDeltaNet, which accounts for pose-dependent deformations in the UV map and is intrinsically learned in our pipeline.
Finally, given an environment map and a virtual camera position, our neural renderer produces the final image of the neural actor.
%
%
\par
To train and evaluate our method, we collected a real-world dataset for novel pose and novel lighting human synthesis, composed of four actors recorded under six different light conditions by a multi-view camera setup, including indoor and outdoor environments and a light probe for each sequence, providing the first real in-the-wild benchmark for controllable human relighting.
%
%
\par 
To summarize, our main technical contributions are as follows: 
1) The first method to learn a relightable neural actor from multi-view video under static lights that allows for pose control, appearance editing, and free-viewpoint rendering under arbitrary illumination while modeling pose-dependent effects.
2) We propose an intrinsic decomposition of diffuse and specular reflection based on a new approach to compute pose-dependent normals and visibility by sparsely sampling a neural field and densifying the results with NormalNet and VisibilityNet, making the problem of pose-driven human relighting tractable. 
Our approach relies on a mapping between 3D and UV space implicitly learned by UVDeltaNet, allowing us to learn an explicit and static material representation that is easy to edit at inference time.
%
%
\par
In our experimental evaluation, we compare our approach with two baselines in the proposed real data benchmark, showing the superiority of our approach in synthesizing relightable humans on challenging motion sequences. The effectiveness of our method is demonstrated by ablation studies and qualitative results.

\section{Related Work}
\label{sec:relatedwork}
Since the goal of this work is to build a \textit{full-body} relightable and animatible avatar, we will not discuss works~\cite{ranjan2023facelit,wang2023sunstage,li2022eyenerf,schwartz2020eyes,deng2023lumigan,tan2022volux,10.2312:EGWR:EGSR07:147-157,yang2023towards,wuu2022multiface,shu2018deforming} focusing on face relighting as both settings have different requirements and challenges.
\subsection{Relighting Static Scenes}
\label{sec:related_work_static}
Classic approaches for relighting and editing perform intrinsic decomposition through inverse rendering~\cite{Ramamoorthi2001_IR}.
With recent advances in neural rendering~\cite{Tewari2022NeuRendSTAR}, light transportation can be taken into account with fully differentiable rendering pipelines.
For instance, Zhang et al.~\cite{zhang2021nerfactor} address the problem of recovering geometry and spatially-varying BRDF with an implicit scene representation. However, this requires solving the light transport by evaluating the NeRF model for each light ray, which is very inefficient. 
NeRV~\cite{Srinivasan2020nerv} and Zhang et al.~\cite{zhang2022modeling} mitigate this problem with a neural visibility field that is jointly optimized with the NeRF model, and NeRD~\cite{boss2021nerd} adds further relaxation by ignoring cast shadows. However, this assumption cannot be made for humans, which are highly articulated and often experience prominent self-casting shadows.
Jin et al.~\cite{jin2023tensoir} propose TensorIR, an efficient tensor-based representation that solves the inverse problem by casting camera and secondary rays. IRON~\cite{zhang2022iron} proposes a two-stage process for obtaining a mesh with material properties, and Lyu et al.~\cite{lyu2022nrtf} and NeRO~\cite{liu2023nero} focus on multi-bounce illumination and reflective objects, which is usually not the case for humans.
The methods mentioned above share the assumption of a static scene and cannot be trivially adapted to dynamic and controllable humans.
%
%
%
%
\subsection{Human Models with Pose Control}
\label{sec:related_work_pose_control}
Modeling humans is a long-standing topic. Traditionally, meshes from an actor can be automatically rigged to a skeleton~\cite{Aguiar2008automatic} using Linear Blending Skinning (LBS)~\cite{Lewis_SIG200_LBS}. Parametric human models offer pose and shape control~\cite{SMPL_2015,SMPL_X_2019} and, more recently, implicit human bodies can be articulated~\cite{COAP_Mihajlovic_CVPR_2022,alldieck2021imghum,mihajlovic2021leap}. Such models offer a generic and controllable representation, but are unable to represent individual details.
Template-based methods synthesize a realistic level of details~\cite{habermann2019livecap, habermann2020deepcap, jiang2022hifecap}, but have to deal with non-linear pose-dependent deformations.
%
%
\par
Several approaches model appearance and geometry with an implicit field anchored on a coarse human mesh. 
Neural Body~\cite{peng2021_neural_body} optimizes latent vectors anchored in the vertices of the human mesh and Neural Actor~\cite{liu2021neural} incorporates texture maps as a conditioning to break down the mapping from pose to dynamic effects (one-to-many mapping).
ARAH~\cite{ARAH_ECCV_2022} models the human surface with an SDF, Dual-Space NeRF~\cite{zhi2022dual} learns the color and light in separate spaces, and Kwon et al.~\cite{kwon2021neural} propose an approach that generalizes to new identities and poses.
Shyshey et al.~\cite{shysheya2019textured} jointly learn textures and a neural renderer but fall short in reconstructing the 3D human body.
A few works~\cite{Weng_2022_CVPR, jiang2022neuman} approach a restrictive scenario of reconstructing 3D humans from a monocular video but fail to model fine details under new poses. 
Common to all these type of methods~\cite{habermann2021, HDHumans2023, kwon2023deliffas, Pang_2024_CVPR} is the entanglement of material properties and illumination, making them restricted to  the same appearance as observed during training. 
\subsection{Relighting Human Models}
\label{sec:related_work_relight_humans}
Only a few methods in the literature tackle the problem of relighting humans from videos under static illumination.
Li et al.~\cite{li2013capturing} create a relightable human mesh by estimating a low-frequency environment using Spherical Harmonics.
Some approaches target the problem of single-image \cite{ji2022geometry} and static pose \cite{zheng2023learning} relighting under unknown lights, which is an extremely ill-posed problem due to the high ambiguity between albedo and lighting.
Relighting4D~\cite{chen2022_Relighting4D} extends Neural Body~\cite{peng2021_neural_body} to estimate geometry and reflectance, but can only replay the same sequence observed during training under new lights and is not able to generate new poses.
RANA~\cite{iqbal2023rana} allows for pose control but assumes Lambertian reflectance, ignoring different material properties such as skin and clothes, and does not model cast shadows, resulting in unrealistic renderings for articulated humans.
%
%
%
In contrast, our method performs intrinsic decomposition with space-varying BRDF, models illumination and self-shadows, allows for pose control and novel lights, and offers an easy manner to edit the material properties of the learned neural actor. 
\section{Method}
\label{sec:method}
Our goal is to learn a digital human avatar (neural actor) from RGB images that models pose-dependent effects and can be relighted, pose-controlled, and rendered from novel viewpoints at test time. 
To achieve relighting, we decompose the observed appearance into normals, visibility with respect to each light source, and BRDF represented as albedo and roughness. 
With these components and an environment map, we employ the microfacet model~\cite{Microfacet_model_EG07} to render the output image of our method, which is compared to the observed images. 
In addition---since we target a drivable actor---the geometry components need to be controllable by the skeletal pose. 
%
%
%
\par
\begin{figure*}[tbp]
    \centering
    \mbox{} \hfill
    \includegraphics[width=1.0\textwidth]{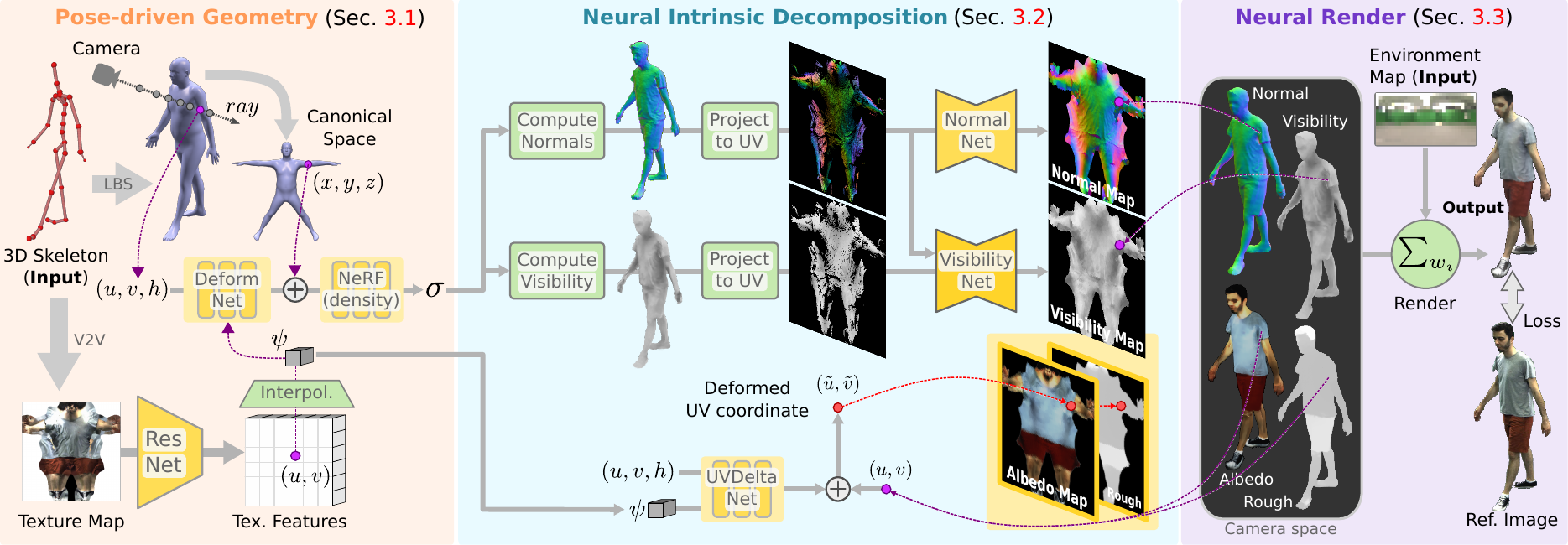}
    \hfill \mbox{}
    \caption{\label{fig:method}%
        Our method takes as input a 3D skeletal pose and a static environment map, and renders the neural actor from a virtual camera position. The pose-driven geometry (\cref{sec:preliminaries}) learns an implicit density function, from which we compute the normal and visibility information. Our intrinsic decomposition disentangles normal, visibility, albedo and roughness maps in UV space (\cref{sec:method_intrinsic_decomposition}). The neural renderer (\cref{sec:method_render}) outputs our prediction, which is supervised with the reference image. \textbf{Yellow} represents learnable components and \textbf{green} non-learnable components. 
    }
\end{figure*}

Fig.~\ref{fig:method} shows an overview of our method divided into three main components. 
The first is a pose-driven implicit geometry model, where the goal is to obtain a skeleton-controllable implicit 3D density field that models the geometry of the actor with pose-dependent deformations. This model is adapted from Neural Actor~\cite{liu2021neural} and is depicted here since our approach builds upon its internal design. Note that Neural Actor does not perform intrinsic decomposition and is unable to perform relighting or appearance editing, which is our main goal.
The second component is our proposed intrinsic decomposition approach whose goal is to disentangle geometry represented as normal and visibility information, and material properties as albedo and roughness. 
Our key insight is to represent these components in UV space rather than in 3D coordinates, using the density field as guidance in the mapping function between 3D and UV spaces. 
This allows our method to model complex deformations that are pose-dependent in a compact UV map that can be easily computed and queried.
Finally, we integrate the microfacet model~\cite{Microfacet_model_EG07} into our rendering pipeline that takes as input the disentangled components (i.e., normal, visibility, albedo, and roughness) and an environment map, and renders the image of the neural actor in an arbitrary pose and light condition from a free-viewpoint. 
\subsection{Preliminaries}  \label{sec:preliminaries}
\subsubsection{Human Pose and Environment Map.}
Our method takes as input a 3D articulated human skeleton pose $\mathbf{P}_t=\{\bm{\theta}_t,\mathbf{R}_t,\mathbf{T}_t\}$, defined by the relative joint rotation  ${\bm{\theta}\in\mathbb{R}^{J\times{3}}}$ of $J$ body joints, the global rotation ${\mathbf{R}\in\mathbb{R}^{3\times{3}}}$, and global translation ${\mathbf{T}\in\mathbb{R}^{3}}$ of the person at frame $t$.
The 3D pose can be used to drive a coarse human mesh with linear blending skinning (LBS)~\cite{SMPL_2015}.
We represent the global lights as an environment map ${\mathbf{L}\in\mathbb{R}^{16\times{32}\times{3}}}$, which can be obtained from HDR images of a mirror sphere~\cite{debevec2008recovering}---see the supplement for more details. The environment map can be static during training and changed during inference to render results under novel illumination.
\par
\subsubsection{Training Data.}
During training, our method is supervised with multi-view calibrated video streams %
$\mathbf{V}_c=\{\mathbf{I}_{c,t}\}$, %
where ${c\in\{1,\dots,C\}}$ is the camera index and $\mathbf{I}_{c,t}$ is the image from camera $c$ at frame $t$.
We obtain a foreground mask $\mathbf{M}_{c,t}$ for each video frame \cite{lin2021real}. 
The coarse human mesh and 3D skeletons can be obtained with multi-view pose tracking~\cite{easymocap}.
The input 3D skeletons drive the LBS model to obtain posed human meshes used to extract texture maps from the multi-view images~\cite{alldieck2018video}.
These texture maps are used during training to learn pose-dependent deformations on the implicit human model.
At inference time, a V2V~\cite{wang2018vid2vid} network predicts texture maps from posed normal maps~\cite{liu2021neural}.
\par
\subsubsection{Implicit Human Model.}
For realistically relighting articulated avatars, we need to model human geometry with deformations caused by dynamic and varied poses.
Therefore, we follow Neural Actor~\cite{liu2021neural} to represent the pose-driven geometry as an implicit field guided by the 3D skeletal pose parameters and conditioned by texture maps, which provide the ability to model high-frequency deformations. 
Specifically, a standard NeRF model~\cite{Mildenhall2020} is optimized through volumetric rendering in the canonical space of the coarse human mesh and the canonical space is deformed based on ResNet~\cite{he2016deep} features extracted from the texture maps, which provide the ability to represent detailed wrinkles and deformations not modeled by the coarse mesh.
During optimization, from each pixel in the reference images we cast a ray ${\mathbf{r}_n=\mathbf{o} + \delta_n\mathbf{d}}$, where $n={1,\dots{N}}$ is the index of each point in the ray, $\mathbf{o}\in\mathbb{R}^3$ is the origin defined in the optical center of the camera, $\mathbf{d}\in\mathbb{R}^3$ is the ray direction, and $\delta_n$ is the depth (distance to the camera).
The sampled points $\mathbf{r}_{n}$ are projected to the mesh surface, resulting in the local coordinates $(u_n,v_n,h_n)$, where $(u,v)$ is the surface coordinate in the UV space and $h$ is the signed distance to the mesh.

%
\subsection{Neural Intrinsic Decomposition} \label{sec:method_intrinsic_decomposition}
At the core of our method, we perform intrinsic decomposition to obtain the normal, visibility, albedo, and roughness components, required for rendering the neural actor under new light conditions.
Specifically, for each point in each ray cast into the scene we want to obtain the normal direction $\mathbf{n}\in\mathbb{R}^3$, the visibility coefficients $\bm{\upsilon}\in\mathbb{R}^{512}$ respective to each light source in the environment map, the RGB albedo $\mathbf{a}\in\mathbb{R}^3$, and the scalar roughness $\rho$.
The normal and visibility values can be computed from the geometry model.
However, computing the visibility of each point in 3D with respect to the light sources is computationally intractable for a dynamic and long sequence, since for every point in 3D and for every frame, several rays have to be cast towards each pixel of the environment map.
Therefore, we propose a different approach. 
\par
We notice that the intrinsic decomposition is only relevant for points close to the surface.
Since the coarse human mesh provides a mapping from the mesh surface to UV space and the implicit human model provides a density field that models the geometry of the actor, we propose an aggregation strategy that samples normal and visibility from a given point of view (therefore, only a few points are sampled in 3D) and aggregates the obtained values in a per-frame UV map.
Once we obtain the normal and visibility in UV space for a given pose, we can simply query the values by projecting the sampled 3D point $\mathbf{r}_n$ into the UV space, preventing our method from the expensive computation of normal and visibility at every possible 3D coordinate over multiple frames.
We next explain this approach in more detail. 
%
%
%
%
\subsubsection{Normal Maps.} \label{sec:method_normals}
For each point in a ray, we want to obtain the normal value that will be required for solving the rendering equation during relighting (will be introduced in \cref{sec:method_render}). 
This value could be obtained by computing the gradient from the neural field~\cite{zhang2021nerfactor}. However, this process requires back-propagating gradients throughout the entire network and often results in noisy normals.
Instead, we explicitly compute the normal vector close to the visible surface by deriving the unbiased depth $\hat{\mathbf{D}}$~\cite{NEURIPS2021_e41e164f}:
\begin{equation}
    \hat{\mathbf{D}}(\mathbf{r})=\frac{\sum_{n=1}^{N}{w_{n}\delta_{n}}}{\sum_{n=1}^{N}{w_{n}}}, \quad \text{with}\;  w_n=T_n\left(1-\text{exp}\left(-\sigma_n{\bar{\delta}_n}\right)\right), 
    \label{eq:unbiased_depth}
\end{equation}
where $T_n=\text{exp}\left(-\sum_{j=1}^{n-1}\sigma_j{\bar{\delta}_j}\right)$, $w_{n}$ is the density weight from volumetric rendering~\cite{Mildenhall2020}, and $\sigma_n$ is the density predicted by the NeRF model at point $\mathbf{r}_n$.

\begin{wrapfigure}{r}{0.5\textwidth}
\begin{center}
\vspace{-2.5em}
    \includegraphics[width=0.49\textwidth]{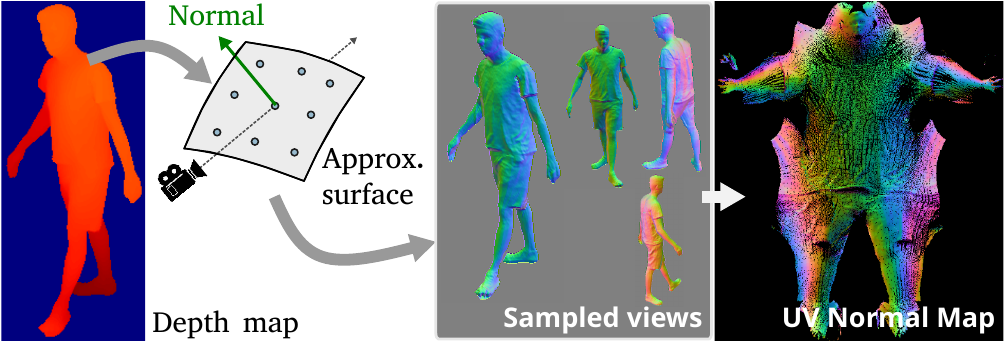}
    \hfill \mbox{}
    \caption{\label{fig:computing_normals}%
      The unbiased depth from \cref{eq:unbiased_depth} is projected into 3D coordinates, where we approximate the tangent surface at the intersecting point and obtain the normal vector. Normals are sampled from multiple viewpoints and aggregated in the UV map. 
    }
    \vspace{-3.0em}
\end{center}
\end{wrapfigure}

Then, as illustrated in \cref{fig:computing_normals}, we convert the depth values relative to the camera to 3D coordinates and obtain the normal direction from the tangent surface computed at the intersection point, defined as the maximum value of the density weight $w_{n}$. This process can be performed for multiple viewpoints to cover the surface of the human.
\par
Once the normal values are obtained in 3D, we project them to the UV map and aggregate them with the obtained normals from different views. 
In this process, the $(u^*,v^*)$ coordinate where the obtained normal will be projected onto the UV map is defined by
\begin{equation}\label{eq:agg_uv_map}
    (u^*, v^*)=\Psi(\mathbf{r}_n^*), \quad \text{with}\;\mathbf{r}_n^*=\argmax_{\mathbf{r}_n} w_n,
\end{equation}
where $\Psi$ is a function that projects a 3D point to the closest surface point in the human mesh. Note that $w_n$ depends on the sampled points $\mathbf{r}_n$ (from $T_n$ in Eq.~\ref{eq:unbiased_depth}) and $(u^*, v^*)$ corresponds to the $(u_n,v_n)$ value obtained from the point $\mathbf{r}_n$ on the ray that has the largest density weight $w_n$.
This process ensures that the normal map obtained aggregates the normals most likely close to the surface. 
%
%
\par
In this process, the resulting UV map can be sparse, which is undesirable for free-viewpoint rendering.
Thus, the NormalNet is used to inpaint the missing normals in UV space. This lightweight CNN is implemented with $8$ layers of partial convolutions~\cite{liu2018partialinpainting}, takes as input sparse normal maps with size $512{\times}512{\times}3$, and predicts a dense normal map (see Fig.~\ref{fig:method}). During training, NormalNet is supervised with densely sampled normal maps from the training data.
\subsubsection{Visibility Maps.} \label{sec:method_visibility}
The visibility information is necessary to model pose-dependent shadows in the neural avatar, without which it is not possible to produce realistic relighting results.
A naive approach for obtaining the visibility for each point in 3D would require querying the neural field thousands of times while casting rays from the intersecting point to all the light sources, which is prohibitive during training. 
In contrast, in our approach we efficiently encode visibility of each point in 3D close to the surface by applying an approximate mapping to UV space.
Specifically, we pre-compute the visibility maps for each pose by sampling multiple camera views and aggregate the results in UV map following \cref{eq:agg_uv_map}.
The resulting visibility maps with size $256\times{256}\times{512}$ encode the visibility information w.r.t. each light source in the environment map (represented by $512$ individual lights).
Similarly to the computation of normal maps, this process results in sparse visibility maps, which are inpainted by the VisibilityNet (as shown in \cref{fig:method}). This shallow CNN takes as input sparse visibility and normal maps, as the visibility highly depends on normals (e.g.~the visibility is always zero for the backlight source), and is trained with densely sampled visibility from the training sequence. See the suplement for archtecture details.
\subsubsection{Albedo and Roughness Maps.} \label{sec:method_albedo_rough}
As discussed above, the normal and visibility components can be computed from the density field by using UV maps as proxies.
However, the albedo and roughness properties cannot be directly obtained from the pre-trained Neural Actor model, since it predicts radiance which is the product of the rendering equation. Instead, we want to learn a decomposed representation.
In intrinsic decomposition, the material properties are commonly optimized via differentiable rendering. 
Hence, we optimize the albedo and roughness of the neural actor by leveraging a differentiable renderer (\cref{sec:method_render}) and by imposing priors through our design.
%
%
%
%
\par
The material properties of an actor should not depend on the pose and should be efficiently encoded to enable fast computations and editing.
Therefore, we model albedo and roughness as static UV maps, respectively defined as ${\mathbf{A}\in\mathbb{R}^{1024\times{1024}\times{3}}}$ and ${\mathbf{B}\in\mathbb{R}^{1024\times{1024}}}$. These tensors are jointly optimized for the whole training sequence and regularized with an L1 loss on the spatial gradients, which encourages smoothness in both maps:
{\small
\begin{equation}
    \mathcal{L}_\mathrm{smooth}=\frac{1}{N}\sum{|\nabla_x(\mathbf{A})|+|\nabla_y(\mathbf{A})|+|\nabla_x(\mathbf{B})|+|\nabla_y(\mathbf{B})|},
    \label{eq:reg_meterial}
\end{equation}
}
where $\nabla_x$ and $\nabla_y$ represent the spatial gradient operators.

Nonetheless, by assuming a material representation that does not depend on the skeletal pose, we still have to cope with intrinsic mesh registration errors.
Such registration errors could cause blurry results or artifacts in the final renderings.
To cope with this problem, we propose a corrective term to the UV mapping, which is predicted by the UVDeltaNet, as shown in \cref{fig:method}.
The UVDeltaNet is an MLP that takes as input the local coordinates $(u,v,h)$ for each point on the ray and the sampled texture features $\psi$, and outputs a corrective term $\mathbf{\epsilon}=(u^\prime, v^\prime)$.
The texture features are necessary to provide additional information, since the local coordinates alone cannot account for deviations in the mesh registration.
%
%
%
%
\par
Initially, UVDeltaNet is frozen and outputs zeros. After the initial convergence of $\mathbf{A}$ and $\mathbf{B}$, UVDeltaNet is indirectly supervised by the rendering loss and regularized with:
\begin{equation}
    \mathcal{L}_\mathrm{uv}=\frac{1}{N}\sum{\bm{\epsilon}^2 + |\nabla_x(\bm{\epsilon})| + |\nabla_y(\bm{\epsilon})|},
    \label{eq:reg_uvdelta}
\end{equation}
which minimizes the magnitude of the deformation and encourages its spatial gradient to be sparse, i.e., it makes the UV deformation spatially coherent.
%
%
%
%
%
%
\subsection{Neural Renderer} \label{sec:method_render} 
The third part of our method is our neural renderer, which takes the disentangled components as normal, visibility, albedo, and roughness; integrates the lights from the environment map, and outputs the rendered image.
%
%
%
%
\par
Our neural renderer starts by recasting the ray $\mathbf{r}$ with importance sampling around the estimated depth value $\hat{\mathbf{D}}$. Thus, all the points $\mathbf{r}_n$ are most likely close to the surface.
For each sampled point in the ray, we obtain the normal direction $\mathbf{n}_n$ and the visibility values $\bm{\upsilon}_n$ by interpolating the normal and visibility maps in UV around the point $(u_n, v_n)=\Psi(\mathbf{r}_n)$.
The albedo $\mathbf{a}_n$ and roughness $\rho_n$ are obtained by interpolating the albedo and roughness maps in UV around the point ${(u_n+u^\prime_u, v_n+v^\prime_n)}$, i.e., following the deformation predicted by UVDeltaNet.
%
%
%
%
\par
Before rendering the final image of the posed neural actor, we consider the light transport equation defined as
\begin{equation}
    \mathbf{L}_0(\mathbf{r}_n, \bm{\omega}_{0})=\sum_{i}\textsc{R}(\mathbf{a}_n,\rho_n,\bm{\omega}_{i},\bm{\omega}_{0})\bm{\upsilon}_{n,i}\mathbf{L}(\bm{\omega}_i)(\bm{\omega}_i{\mathbf{n}_n}),
    \label{eq:light_transport}
\end{equation}
where $\textsc{R}(\cdot)$ is the microfacet BRDF model~\cite{Microfacet_model_EG07}, $\bm{\omega}_{0}$ and $\bm{\omega}_{i}$ are the camera and the light source directions, respectively, $\bm{\upsilon}_{n,i}$ is the visibility of point $\mathbf{r}_n$ w.r.t. the i$th$ light source, and $\mathbf{L}(\bm{\omega}_i)$ is the incoming light from the environment map.
\cref{eq:light_transport} is defined for a single point $\mathbf{r}_n$ on the ray and is integrated into our volume rendering as 
\begin{equation}
    \tilde{\mathbf{I}}(\mathbf{r})=\sum_{n=1}^{N}{w_n}\mathbf{L}_0(\mathbf{r}_n, \bm{\omega}_{0}),
    \label{eq:relight_image}
\end{equation}
where $\tilde{\mathbf{I}}$ is the output image of the relighted neural actor. 
%
%
%
%
\par
During training, our relightable model is supervised with the following loss:
\begin{equation}
    \mathcal{L}_{\text{relight}}=\lambda_{\text{L2}}\mathcal{L}_{\text{L2}}%
        +\lambda_{\text{vgg}}\mathcal{L}_{\text{vgg}}%
        +\lambda_{\text{smooth}}\mathcal{L}_{\text{smooth}}%
        +\mathcal{L}_\mathrm{uv},
\end{equation}
where $\mathcal{L}_{\text{L2}}$ and $\mathcal{L}_{\text{vgg}}$ are the L2 and the perceptual losses \cite{vgg_loss_2017} between $\tilde{\mathbf{I}}(\mathbf{r})$ and the reference image. The hyperparameters are defined as $\lambda_{\text{L2}}=100$, $\lambda_{\text{vgg}}=0.01$, and $\lambda_{\text{smooth}}=0.1$ in all our experiments.

\section{Relightable Dynamic Actors Dataset}
\label{sec:dataset}
\begin{figure*}[tbpt]
    \centering
    \mbox{} \hfill
    \includegraphics[width=1.0\textwidth]{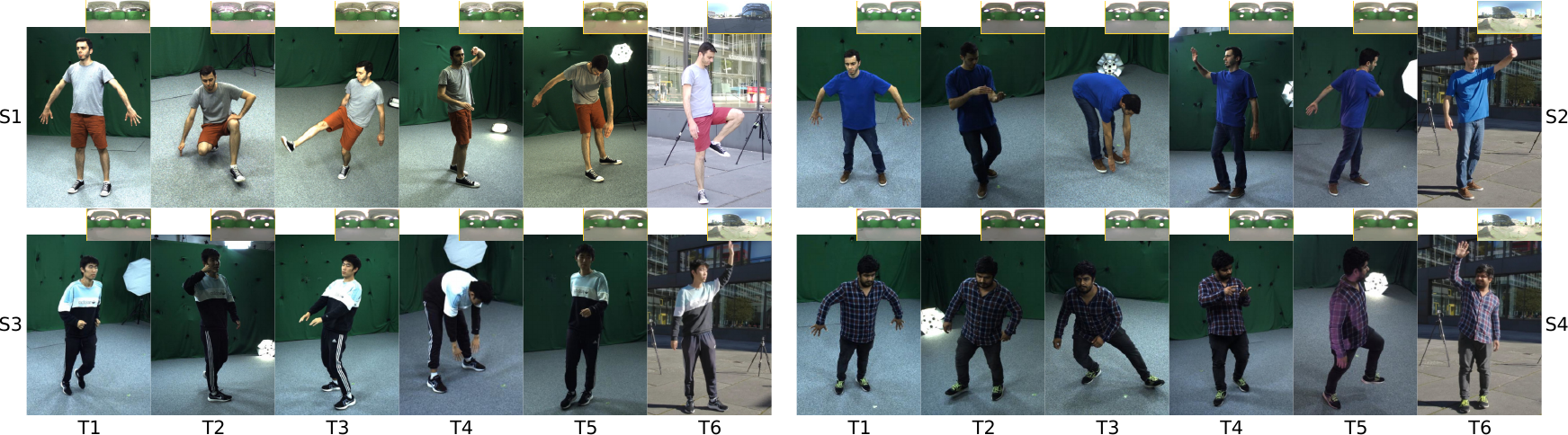}
    \hfill \mbox{}
    \caption{\label{fig:dataset_samples}%
        \textit{Relightable Dynamic Actors} dataset. For each actor, we recorded six sequences under different light conditions, including five indoor sequences and one outdoor. Environment maps were obtained from HDR pictures of a mirror sphere.
    }
\end{figure*}
Existing datasets for human reenactment and relighting are often monocular~\cite{alldieck2018video, iqbal2023rana}, contain very short sequences with only a few pose variations, and do not provide environment maps~\cite{liu2021neural, peng2021_neural_body, alldieck2018video, iqbal2023rana}. 
Hence, we cannot use them for evaluation under realistic scenarios with real ground truth. 
In addition, we observe that the only existing synthetic dataset for human relighting and pose control~\cite{iqbal2023rana} suffers from artifacts in the geometry and salt-and-pepper noise in the renderings. 
With these shortcomings in mind, we collect a new real dataset, which we call \textit{Relightable Dynamic Actors}, with four identities performing a series of ten different actions (Fig.~\ref{fig:dataset_samples}). 
Each subject is recorded six times wearing the same clothes but under different light conditions, including indoor and outdoor, where we used around $100$ and eight cameras, respectively. 
For each sequence, we also capture HDR light probes, which can be used for training or testing. 
In total, our dataset contains around $90$k multi-view video frames divided into $24$ multi-view video sequences. 
We refer to the supplemental document for more details. 
\section{Experiments}
\label{sec:experiments}
In this section, we provide an experimental evaluation of our method on the Relightable Dynamic Actors dataset and show qualitative and quantitative results. Please refer to the supplementary video for dynamic results on long sequences.
\subsection{Evaluation}
\par
\noindent
\textbf{Metrics}.
For quantitative evaluation, we report results on the PSNR, SSIM and LPIPS metrics~\cite{zhang2018perceptual}, comparing the rendered relight results with the ground-truth images from our dataset.
Importantly, our work is the first to be able to report such metrics on a real dataset since no previous dataset in the literature provided a real human recorded under different and known light conditions.
%
%
%
\par
\noindent
\textbf{Baselines}.
Since the only method in the literature capable of reposing a human under new light conditions does not provide implementation~\cite{iqbal2023rana}, and Relighting4D \cite{chen2022_Relighting4D} is not able to perform inference for new poses, we defined two baselines for a fair comparison.
Our baselines are derived from Neural Actor~\cite{liu2021neural} and adapted to perform intrinsic decomposition based on the obtained normal and visibility information, as discussed in Sec.~\ref{sec:method_normals}.
However, instead of using NormalNet and VisibilityNet to obtain these components, we densely sample the neural field and perform morphological inpainting to obtain dense UV maps.
The first baseline, referred to as \textit{Neural Actor+IR}, predicts the albedo and roughness components instead of RGB color directly from the deformed canonical space $\mathbf{x}_{n}=(x,y,z)$ (see \cref{fig:method}). The second baseline, \textit{Neural Actor+IR+Tex}, takes as input the texture features $\psi$ as a conditioning for predicting the albedo and roughness values. In both cases, the predictor is a NeRF MLP with 8 layers.
\par
\noindent
\textbf{Implementation Details}.
Our model is trained in a three-stage process. First, we train the geometry component (\cref{sec:preliminaries}) with the Adam~\cite{kingma2014adam} optimizer with a learning rate of $0.001$ until convergence, which takes about two days on four NVidia A40 GPUs. Then, we pre-train the NormalNet and VisibilityNet with pre-computed normal and visibility UV maps sampled from the geometry model from $40$ different views. Both networks are trained in parallel using Adam with a learning rate of $0.01$ for about $12$ hours. Finally, we train our intrinsic decomposition model with the same optimizer until convergence, which takes about one day.
Since the obtained environment maps are up to factor of scale~\cite{debevec2008recovering}, we apply global color correction during training and evaluation.
The predicted texture maps are obtained with a V2V model~\cite{wang2018vid2vid} trained (in parallel to our model) to translate a sequence of SMPL normal maps into a sequence of texture maps.
Please refer to our supplementary materials for architecture details.

\subsection{Results}
\begin{figure}[tbp]
    \centering
    \mbox{} \hfill
    \includegraphics[width=0.75\textwidth]{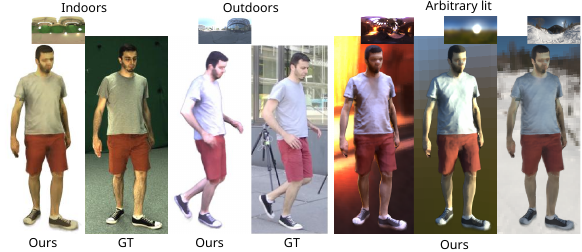}
    \hfill \mbox{}
    \vspace{-1em}
    \caption{\label{fig:qualitative_results}%
        Qualitative results of our method on \textit{novel poses}. Our approach generalizes to both indoor and outdoor illumination, and generates realistic renderings for arbitrary environment maps with detailed shadows and pose-dependent deformations.
    }
\end{figure}
\noindent
\textbf{Qualitative Results}.
In \cref{fig:qualitative_results}, we show qualitative results of our method. Our results are all from a trained model driven by a \textit{new skeletal pose}, \textit{new camera viewpoint}, and \textit{new environment map} not seen during training. Note how our approach produces realistic appearance and prominent cloth wrinkles, which can be easily observed under different light conditions.
Importantly, notice how our results look realistic under novel lights, even though our model is trained on a single sequence under a static light condition. 
\begin{figure}[tbp]
    \centering
    \mbox{} \hfill
    \includegraphics[width=0.75\textwidth]{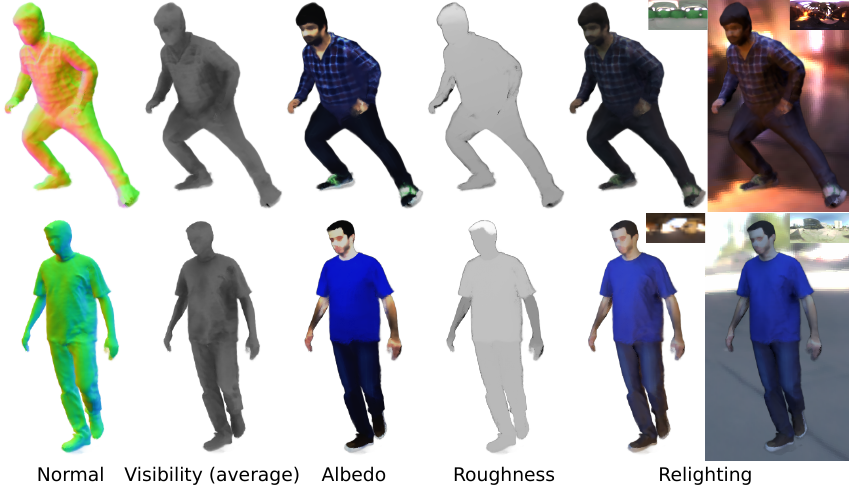}
    \hfill \mbox{}
    \vspace{-1em}
    \caption{\label{fig:intrinsic_decomposition}%
        Results from our intrinsic decomposition. Our method provides detailed normals and visibility (averaged for visualization), and plausible albedo and roughness predictions that enable realistic relighting.
    }
\end{figure}
\begin{table*}[htp]
    \centering
    \caption{\label{tab:quantitative_results}%
      Quantitative results from our method on real data, subject S1, considering indoor and outdoor scenes. Our method consistently improves the baselines, especially in outdoor scenarios, where the gap between training and test lights is more pronounced.
    }
    \begin{adjustbox}{width=1.\textwidth,center}
    \begin{tabular}{@{}lccc|ccc|ccc|ccc|ccc@{}}
    \toprule
    \multirow{2}{*}{Method} & \multicolumn{3}{c}{T2} & \multicolumn{3}{c}{T3} & \multicolumn{3}{c}{T4} & \multicolumn{3}{c}{T5} & \multicolumn{3}{c}{T6 (outdoors)} \\ \cmidrule(l){2-16} 
                 & PSNR~$\uparrow$ & SSIM~$\uparrow$     & LPIPS~$\downarrow$   & PSNR~$\uparrow$     & SSIM~$\uparrow$     & LPIPS~$\downarrow$ & PSNR~$\uparrow$     & SSIM~$\uparrow$     & LPIPS~$\downarrow$  & PSNR~$\uparrow$        & SSIM~$\uparrow$       & LPIPS~$\downarrow$       & PSNR~$\uparrow$        & SSIM~$\uparrow$       & LPIPS~$\downarrow$       \\ \midrule
    Neural Actor+IR     & 17.522 & 0.750 & 0.181 & 17.473 & 0.768 & 0.178 & 17.745 & 0.767 & 0.171 & 18.244 & 0.768 & 0.175 & 18.355   & 0.710  & 0.203 \\
    Neural Actor+IR+Tex  & 18.175 & 0.778 & 0.164 & 17.799 & 0.790 & \textbf{0.167} & 18.132 & 0.785 & 0.167 & 18.531 & 0.789 & 0.166 & 18.704 & 0.737 & 0.186 \\
    \textbf{Ours}        & \textbf{18.602} & \textbf{0.792} & \textbf{0.163} & \textbf{18.240} & \textbf{0.800} & 0.169 & \textbf{18.388} & \textbf{0.800} & \textbf{0.164} & \textbf{18.820} & \textbf{0.800} & \textbf{0.165} & \textbf{19.461} & \textbf{0.753} & \textbf{0.173} \\ \bottomrule
    \end{tabular}
    \end{adjustbox}
    \vspace{-1.0em}
\end{table*}
\par
In \cref{fig:intrinsic_decomposition}, we show results from our intrinsic decomposition. Note that our normal representation nicely captures the geometry details and the visibility. This is important to obtain an albedo that does not contain baked-in lighting effects, e.g. shadows, as can be observed from our results.
\begin{wrapfigure}{r}{0.4\textwidth}
\begin{center}
\vspace{-4.0em}
    \mbox{} \hfill
    \includegraphics[width=0.4\textwidth]{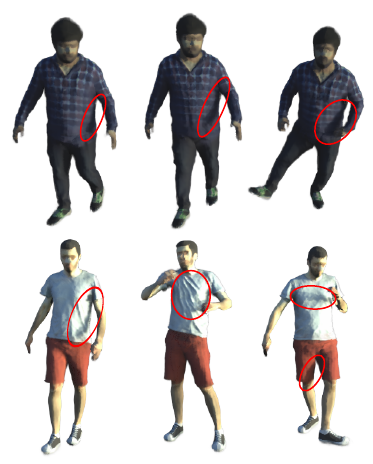}
    \hfill \mbox{}
    \vspace{-1em}
    \caption{\label{fig:cast_shadows}%
        Our method models pose-dependent effects and cast shadows, as can be observed from our results under the ``sunrise'' light.
    }
    \vspace{-3.0em}
\end{center}
\end{wrapfigure}
Another important and unique characteristic of our method is the ability to model self-shadows and cloth wrinkles. In \cref{fig:cast_shadows}, we show results from two subjects, where a ``sunrise'' light probe was used to produce cast shadows. Note how pose-dependent effects are consistent with the mostly directional illumination.
%

%
%
%
\par
\noindent
\textbf{Comparisons}.
We evaluated our method on the test sequences from our dataset. Specifically, we compare against two baselines in the indoor sequences (T2-T5) and outdoor sequence (T6) of subject S1 in every $10$th frame, which sums up to $4000$ evaluation frames. In \cref{tab:quantitative_results} we show, for the first time, quantitative results of \textit{real human relighting} under new poses and new light conditions. Our approach consistently improves upon the two baselines.
%
%
%
%
\par
\noindent
\textbf{Ablation}.
In \cref{tab:ablation}, we show an ablation with different components of our method.
Compared to our baseline (Neural Actor+IR), learning the albedo and roughness as UV maps guided by the neural field significantly improves the performance on relighting and reposing. This is because, in the baselines, the albedo and roughness are under-constrained and can possibly embed shading effects, overfitting to the training data.
The NormalNet and VisiblityNet provide additional capacity to the model to learn how to inpaint the sampled UV maps. Note that without these two models our results are computed using morphological region growth in the UV maps.
Finally, the UVDeltaNet provides more flexibility to the albedo and roughness representation, accounting for fine details. This can be observed in the improvement in the LPIPS metric in the last row of \cref{tab:ablation}. Nonetheless, this also means more capacity to overfit the training sequences, which translates into a small decrease in the PSNR metric.

\begin{wraptable}{r}{0.6\textwidth}
  \begin{center}
  \vspace{-7em}
    \caption{\label{tab:ablation}%
      Ablation study with different components of our method. The proposed intrinsic decomposition with static albedo and roughness maps significantly improves over the baseline, and the proposed NormalNet and VisibilityNet further improve the results while enabling free-viewpoint rendering. UVDeltaNet provides more realistic results (LPIPS metric) with a small sacrifice in the PSNR.
    }
    \begin{adjustbox}{width=0.59\textwidth,center}
    \begin{tabular}{@{}lcccccc@{}}
    \toprule
    \multirow{2}{*}{Method} & \multicolumn{3}{c}{Indoors (S1,T2)} & \multicolumn{3}{c}{Outdoors (S1,T6)} \\ \cmidrule(l){2-7} 
                       & PSNR~$\uparrow$ & SSIM~$\uparrow$ & LPIPS~${\downarrow}$ & PSNR~$\uparrow$ & SSIM~$\uparrow$ & LPIPS~${\downarrow}$ \\ \midrule
    Baseline           & 17.522 & 0.750 & 0.181 & 18.355 & 0.710 & 0.203 \\ 
    +Albedo/Rough. Maps & 18.334 & 0.789 & 0.168 & 19.341 & 0.747 & 0.181 \\ 
    +Normal/Vis.Net    & \textbf{18.634} & \textbf{0.792} & 0.173 & \textbf{19.704} & \textbf{0.760} & 0.180 \\ 
    +UVDeltaNet (\textbf{Ours}) & 18.602 & \textbf{0.792} & \textbf{0.163} & 19.461 & 0.753 & \textbf{0.173}  \\ \bottomrule
    \end{tabular}
    \end{adjustbox}
    \vspace{-3em}
\end{center}
\end{wraptable}
\subsection{Applications}
\begin{wrapfigure}{r}{0.5\textwidth}
\begin{center}
\vspace{-6em}
    \mbox{} \hfill
    \includegraphics[width=0.49\textwidth]{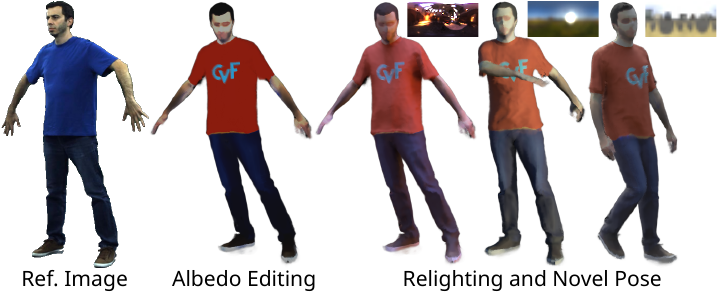}
    \hfill \mbox{}
    \vspace{-1em}
    \caption{\label{fig:applications}%
        Our method allows editing the material properties and simultaneously performing skeletal pose control, free-viewpoint rendering, and relighting of the edited neural actor.
    }
    \vspace{-3em}
\end{center}
\end{wrapfigure}

An important characteristic of our approach is our static representation in UV for the albedo and roughness. This allows exciting and new applications such as color and material editing of an animatable and relightable neural actor, as shown in \cref{fig:applications} and in \cref{fig:teaser}. With a simple edit in the obtained albedo or roughness 2D map, we can insert new textures or new materials and render the edited neural actor under new poses, new viewpoints, and new light conditions.

\section{Limitations}
\label{sec:limitations}
This work focuses on visual fidelity, and representations faster than NeRF could be tested in our framework in future. 
Moreover, like several previous techniques for the free-viewpoint rendering of humans and avatar generation \cite{liu2021neural, Remelli2022}, the proposed method can currently only deal with tight clothing due to the coarse geometry registration. The coarse mesh registration can also negatively affect the fine details in the face and hand synthesis.
\section{Conclusion}
\label{sec:conclusion}
We conclude that the proposed method for the free-viewpoint rendering of humans enables simultaneous relighting, material editing, and pose control at test time while requiring a single multi-view video sequence with static illumination during training.
Our method demonstrates functionality that no competing method making the same assumptions at training time (i.e.~from multi-view RGB video inputs only) can provide. 
We hope that our method and the proposed dataset will provoke further research in this exciting field, bringing the neural rendering of humans to a new qualitative level and paving the way for next-generation metaverse applications. 
\section*{Acknowledgements}
This work was funded by the ERC Consolidator Grant 4DRepLy (770784), the Saarbrücken Research Center for Visual Computing, Interaction and AI, and the German Research Foundation (DFG) under Grant No. 468670075.

%
%
\bibliographystyle{splncs04}
\bibliography{references}

\newpage

\title{Relightable Neural Actor with Intrinsic Decomposition and Pose Control \\
---Supplementary Material---
}
\titlerunning{Relightable Neural Actor -- Supplementary Material}
\authorrunning{D.C.~Luvizon et al.}

\author{}
\institute{}

\clearpage
\maketitle

This document accompanies our paper ``Relightable Neural Actor with Intrinsic Decomposition and Pose Control'' and includes additional implementation details on network architectures (\cref{sec:supp_network_arch}), dataset collection (\cref{sec:supp_data_capture}), and additional results and comparisons of our method (\cref{sec:additional_results}).
Our dynamic results are provided in the supplementary video.
\section{Implementation Details and Network Architectures} \label{sec:supp_network_arch}
\subsection{Pose-driven Geometry Model}
\begin{figure}[hbpt]
    \centering
    \mbox{} \hfill
    \includegraphics[width=0.9\textwidth]{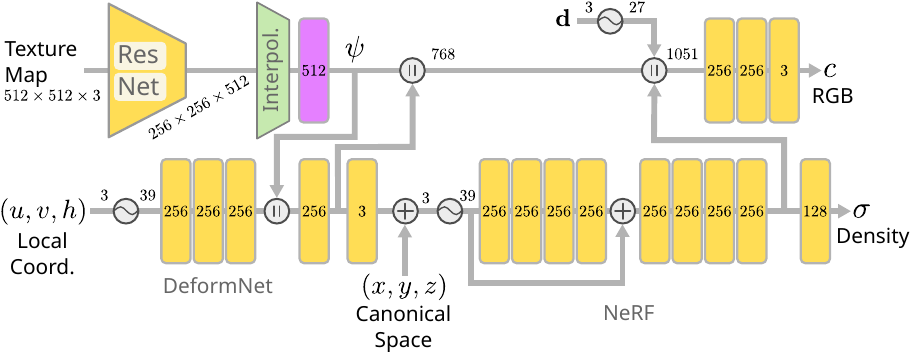}
    \hfill \mbox{}
    \caption{\label{fig:suppmat_arch_geometry}%
        Neural network architecture of the pose-driven geometry model adapted from Neural Actor~\cite{liu2021neural}. The RGB color branch is only used during the training phase of the geometry model and is discarded during the relighting training and inference. The interpolation of texture features (obtained from the texture map) is performed in the UV coordinate $(u,v)$ given by the projection of 3D points onto the human mesh. The obtained feature vector $\psi$ (represented in purple) is also used in our UVDeltaNet. The symbol ``$\sim$''refers to positional encoding as in NeRF~\cite{Mildenhall2020} and ``$||$'' refers to feature-wise concatenation.
    }
\end{figure}

We represent the human geometry as an implicit field guided by the skeletal pose parameters $\mathbf{P}_t$. We follow the architecture from Neural Actor~\cite{liu2021neural}. However, differently from the prior work, we sample image patches during training, allowing our method to process only the triangles that intersect the patch region and enabling the use of perceptual loss during training, as described in Section 3.3 of the main paper.
For completeness, we illustrate the neural network architecture of the pose-driven geometry model in \cref{fig:suppmat_arch_geometry}. The RGB color output $\mathbf{c}$ is activated with the \textit{sigmoid} function, and the density output $\sigma$ is activated with ReLU. The yellow rectangular blocks represent a single fully connected layer.
%
%
%
%
\par
Since our goal at this stage is to obtain a drivable high-quality geometry, we only use the RGB color output $\mathbf{c}$ for supervising the geometry model.
The loss for this part is defined as:
\begin{equation}
    \mathcal{L}_{\text{geo}}=\lambda_{\text{L2}}\mathcal{L}_{\text{L2}}%
        +\lambda_{\text{vgg}}\mathcal{L}_{\text{vgg}}%
        +\lambda_{\sigma}\mathcal{L}_\sigma,
\end{equation}
where $\mathcal{L}_{\text{L2}}$ and $\mathcal{L}_{\text{vgg}}$ are the L2 and perceptual losses ~\cite{vgg_loss_2017} between the ground-truth and the predicted images, and $\mathcal{L}_\sigma$ pushes the density in the empty space to zero. 
Specifically, the density defined as $\sigma=\text{ReLU} (\sigma^\prime)$ suffers from zero-gradient on empty regions and the pre-activated $\sigma^\prime$ is pushed towards negative values with 
\begin{equation}
    \mathcal{L}_\sigma=\left(1-\mathbf{M}(\mathbf{r})\right)\text{sigmoid}(\sigma^\prime),
\end{equation}
where $\mathbf{M}(\mathbf{r})$ is the binary mask at the pixel intersected by the ray~$\mathbf{r}$. The coefficients $\lambda_{(.)}$ are defined empirically and set to $\lambda_{\text{L2}}=100$ and $\lambda_{\text{vgg}}{=}\lambda_{\sigma}{=}0.01$ in all experiments.
After training, we discard the RGB color output and only use the density $\sigma$ to implicitly represent the geometry of the neural actor.
\subsection{NormalNet Model}
\begin{figure}[hbpt]
    \centering
    \mbox{} \hfill
    \includegraphics[width=0.8\textwidth]{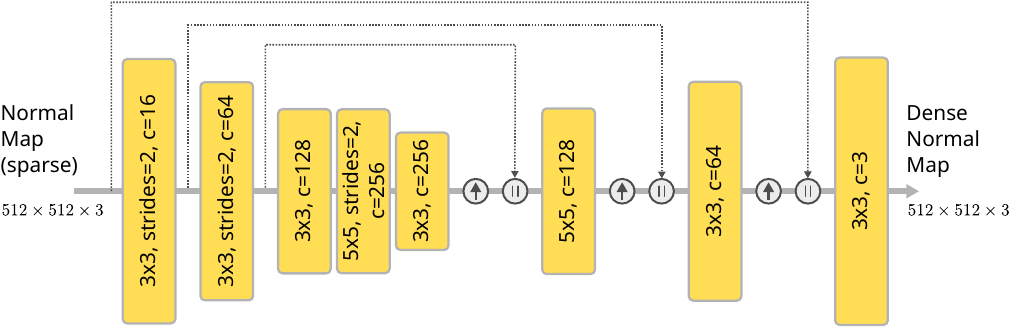}
    \hfill \mbox{}
    \caption{\label{fig:suppmat_arch_normalnet}%
        Neural network architecture of our NormalNet model. Each rectangular block is a 2D partial convolution. The symbol ``$\uparrow$'' refers to depth-to-space transformation, where the spatial resolution is increased by a factor of $2$ in each dimension, and ``$||$'' refers to feature-wise concatenation.
    }
\end{figure}

The goal of NormalNet in our approach is to perform inpainting and refinement of the normal values sampled from the implicit neural field and aggregated in the UV map. Therefore, we leverage partial convolutions~\cite{liu2018partialinpainting} as a building block and design a shallow convolutional neural network (CNN) that is illustrated in \cref{fig:suppmat_arch_normalnet}. 
\subsection{VisibilityNet Model}
\begin{figure}[thbp]
    \centering
    \mbox{} \hfill
    \includegraphics[width=0.8\textwidth]{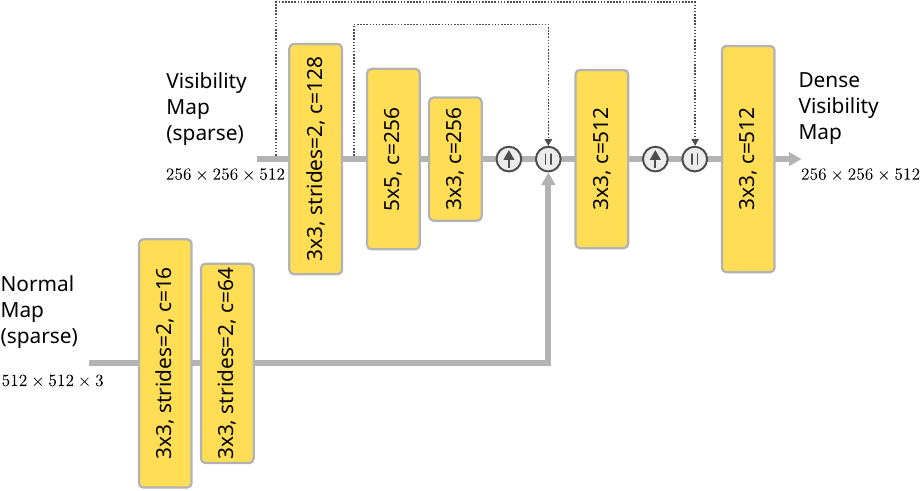}
    \hfill \mbox{}
    \caption{\label{fig:suppmat_arch_visibilitynet}%
        Neural network architecture of our VisibilityNet model. Rectangular blocks represent 2D partial convolutions. ``$\uparrow$'' and ``$||$'' refer to depth-to-space transformation and feature-wise concatenation, respectively.
    }
\end{figure}

We use a similar approach as in the NormalNet architecture for inpainting and refining the visibility information. However, the visibility is highly dependent on the normal values, since half of the light sources in the environment map are back lit as a function of the normal direction. Therefore, the VisibilityNet model also takes as input the sampled normal maps. The shallow CNN architecture of VisibilityNet is shown in \cref{fig:suppmat_arch_visibilitynet}.
\subsection{UVDeltaNet Model}
\begin{figure}[hbpt]
    \centering
    \mbox{} \hfill
    \includegraphics[width=0.55\textwidth]{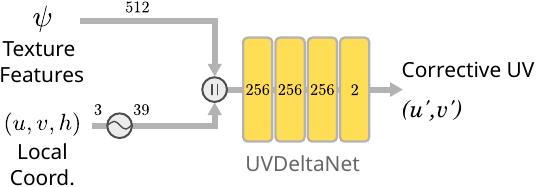}
    \hfill \mbox{}
    \caption{\label{fig:suppmat_arch_uvdeltanet}%
        Neural network architecture of our UVDeltaNet model. For predicting the corrective UV term, this model takes as input the local coordinates $(u,v,h)$, corresponding to the 3D sampled point projected onto the human mesh surface, and the ResNet texture features $\psi$.
    }
\end{figure}

The architecture of UVDeltaNet is shown in \cref{fig:suppmat_arch_uvdeltanet}. This shallow MLP has as objective predict a corrective term in the UV space, therefore it takes as input the local coordinates $(u,v,h)$ and the texture features $\psi$ (see \cref{fig:suppmat_arch_geometry}), and predicts a corrective term $(u^\prime, v^\prime)$.
\begin{figure}[tbp]
    \centering
    \mbox{} \hfill
    \includegraphics[width=1.0\textwidth]{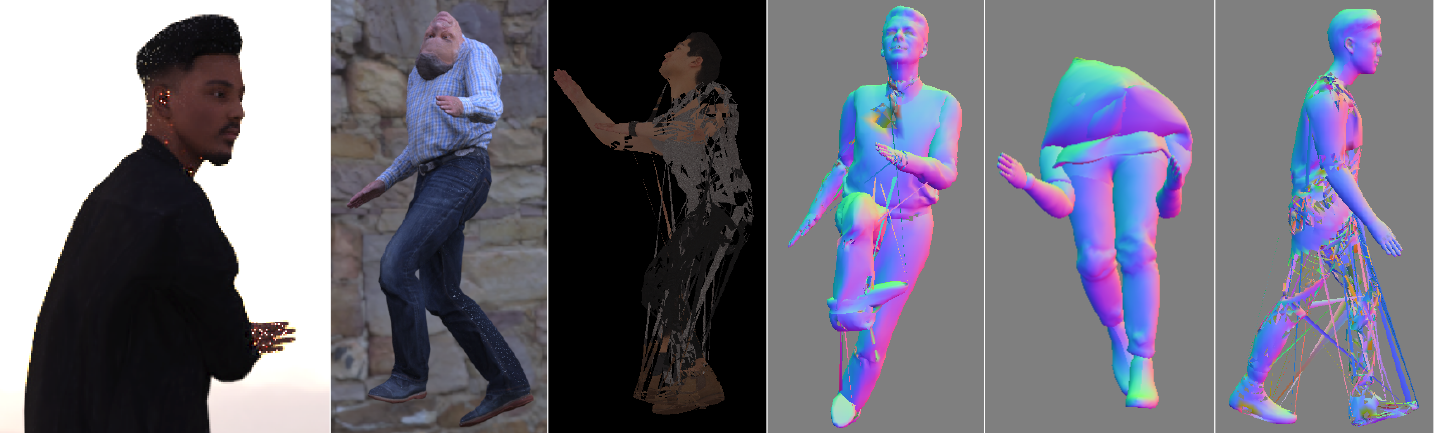}
    \hfill \mbox{}
    \caption{\label{fig:dataset_samples_RANA}%
        Samples from the provided synthetic dataset in RANA~\cite{iqbal2023rana}. Rendered images (left) present salt and pepper noise, as well as severe geometry artifacts (right).
    }
\end{figure}

\section{Dataset Collection} \label{sec:supp_data_capture}
The only existing dataset for novel pose human relighting with ground-truth novel poses under new light conditions is proposed in RANA~\cite{iqbal2023rana}. However, as can be seen in \cref{fig:dataset_samples_RANA}, the provided synthetic data presents severe image and geometry artifacts, despite being limited to a small number of frames from a monocular view of the person.
\par
Therefore, we collected the new dataset ``Relightable Dynamic Actors'' with the goal of providing a real-world dataset with the same person recorded under different light conditions.
\begin{figure}[hbp]
    \centering
    \mbox{} \hfill
    \includegraphics[width=0.65\textwidth]{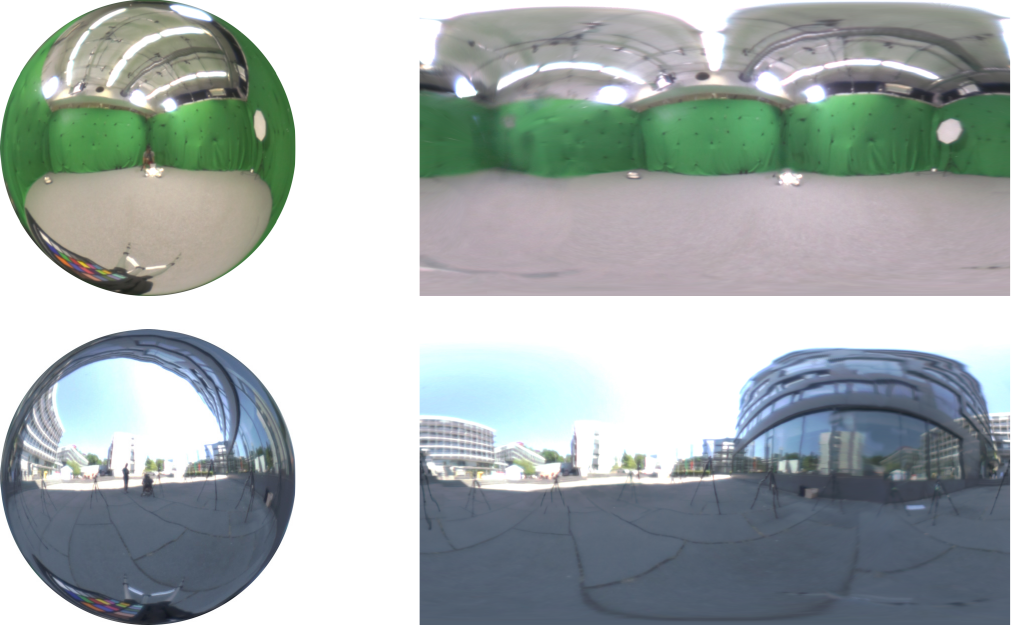}
    \hfill \mbox{}
    \caption{\label{fig:env_maps}%
        Samples from our environment maps obtained from indoor and outdoor sequences. On the left is the light probe obtained with HDR pictures from a mirror sphere and on the right is the processed HDR images converted to a latitude-longitude format (EXR). The environment maps (right) are then resized to $32\times{16}$ pixels to be used in our method.
    }
\end{figure}
Please refer to the main paper for samples from our data.
For each sequence in our dataset, the actors were instructed to casually follow a sequence of $10$ activities defined as: \textit{stretch arms}, \textit{walk in a circle}, \textit{jogging}, \textit{stretch legs}, \textit{talk on the phone}, \textit{use a tablet}, \textit{stretch legs up}, \textit{stretch back}, \textit{wave hands}, \textit{freestyle}.
We use a commercial markerless motion tracking system~\cite{captury} for obtaining the human motion in 3D from the calibrated multi-view videos. To fit SMPL model~\cite{SMPL_2015} to our tracking, we first optimize the pose and shape parameters in the rest pose ``T-pose'' with a 3D body joint loss between SMPL and the tracked 3D pose based on EasyMocap~\cite{easymocap}, then track the SMPL model following the ground-truth motion.
\par
We obtained the environment map for each sequence by taking multi-exposure pictures from a mirror sphere. The HDR image of the mirror sphere was obtained with the algorithm from Debevec et al.~\cite{debevec2008recovering}. We combined two shots of the mirror sphere from different angles. The combined image was unwrapped to the EXR latitude-longitude format, resulting in the HDR environment map. Two sampled from our environment maps are shown in \cref{fig:env_maps}.

\begin{figure}[htbp]
    \centering
    \mbox{} \hfill
    \includegraphics[width=0.99\textwidth]{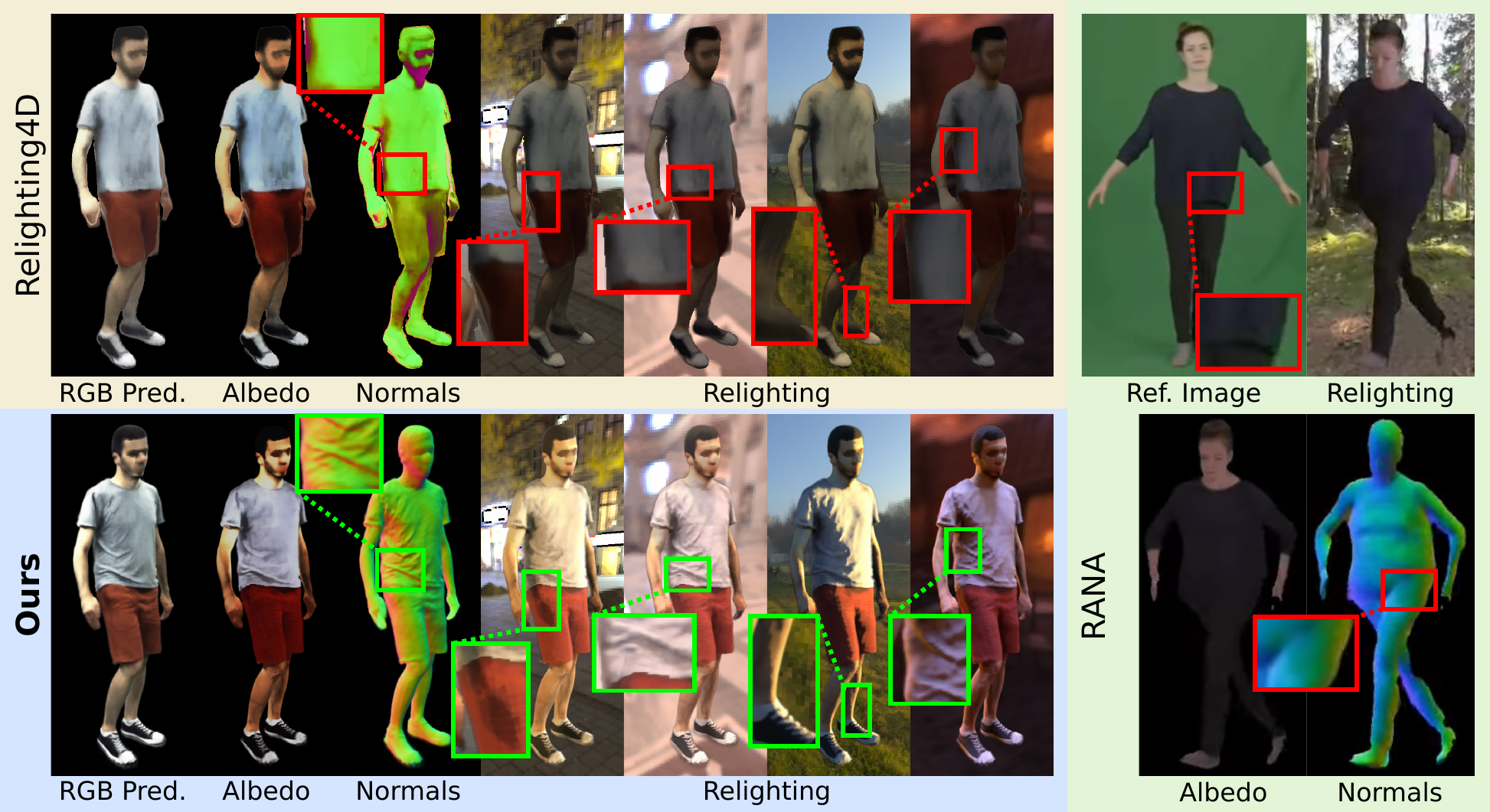}
    \hfill \mbox{}
    \caption{\label{fig:comparisons_r4d_rana}%
        Comparison between our method, Relighting4D~\cite{chen2022_Relighting4D}, and RANA~\cite{iqbal2023rana}.
    Note how Relighting4D and RANA fail to recover fine details on the surface and produce unrealistic results.
    Our results are much sharper and with realistic shading and colors under new challenging lights that produce strong cast shadows.
    }
\end{figure}

\begin{figure}[htbp]
    \centering
    \mbox{} \hfill
    \includegraphics[width=0.9\textwidth]{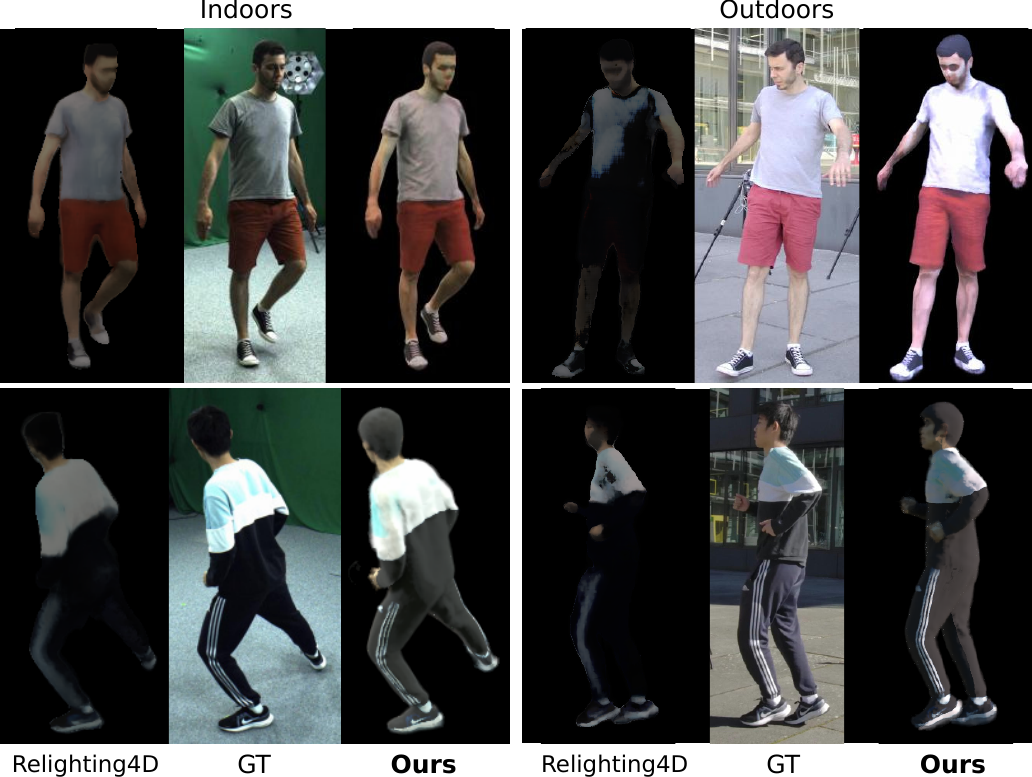}
    \hfill \mbox{}
    \caption{\label{fig:comparisons_r4d}%
        Qualitative comparisons between Relighting4D~\cite{chen2022_Relighting4D} and our method. Our results capture more fine details such as clothe wrinkles and shadows, while Relighting4D produces blurry results with severe artifacts due to its inability to synthesize realistic renderings under novel poses.
    }
\end{figure}

\section{Additional Results and Comparisons} \label{sec:additional_results}

Comparisons between the quality of our method and existing approaches are shown in Fig.~\ref{fig:comparisons_r4d_rana}.
We further compare our method with Relighting4D~\cite{chen2022_Relighting4D}, as shown in \cref{tab:comparison_r4d}.
Since Relighting4D can only replay the same sequence under new light conditions, we adapted it to render the virtual human under the new poses from our test sequences using the respective ground-truth environment maps. Relighting4D was trained using the same setup as our method, i.e., multi-view video sequence with all the training camera views.

\begin{table*}[htp]
    \centering
    \caption{\label{tab:comparison_r4d}%
      We compared our method with Relighting4D~\cite{chen2022_Relighting4D} on our real data, subject S1, considering indoor and outdoor scenes.
      Relighting4D was originally designed to replay the same training sequence. Therefore, we adapted it in this experiment to produce the new poses from our test sequences under new light conditions.
      Our approach consistently outperforms under novel poses, specially under outdoors illumination, which strongly differs from the training lights.
    }
    \begin{adjustbox}{width=1.\textwidth,center}
    \begin{tabular}{@{}lccc|ccc|ccc|ccc|ccc@{}}
    \toprule
    \multirow{2}{*}{Method} & \multicolumn{3}{c}{T2} & \multicolumn{3}{c}{T3} & \multicolumn{3}{c}{T4} & \multicolumn{3}{c}{T5} & \multicolumn{3}{c}{T6 (outdoors)} \\ \cmidrule(l){2-16} 
                 & PSNR~$\uparrow$ & SSIM~$\uparrow$     & LPIPS~$\downarrow$   & PSNR~$\uparrow$     & SSIM~$\uparrow$     & LPIPS~$\downarrow$ & PSNR~$\uparrow$     & SSIM~$\uparrow$     & LPIPS~$\downarrow$  & PSNR~$\uparrow$        & SSIM~$\uparrow$       & LPIPS~$\downarrow$       & PSNR~$\uparrow$        & SSIM~$\uparrow$       & LPIPS~$\downarrow$       \\ \midrule
    Relighting4D~\cite{chen2022_Relighting4D}  & 16.108 & 0.743 & 0.183 & 15.728 & 0.737 & 0.171 & 16.969 & 0.748 & 0.172 & 16.759 & 0.739 & 0.166 & 8.492 & 0.600 & 0.296 \\
    \textbf{Ours}    & \textbf{18.602} & \textbf{0.792} & \textbf{0.163} & \textbf{18.240} & \textbf{0.800} & \textbf{0.169} & \textbf{18.388}  & \textbf{0.800} & \textbf{0.164} & \textbf{18.820} & \textbf{0.800} & \textbf{0.165} & \textbf{19.461} & \textbf{0.753} & \textbf{0.173} \\ \bottomrule
    \end{tabular}
    \end{adjustbox}
\end{table*}

In \cref{fig:comparisons_r4d} we show qualitative comparisons between Relighting4D and our method.  Note that Relighting4D is enable to produce realistic results under our settings, i.e., novel pose and novel light conditions, even when it is trained on the same setup. The superiority of our approach can be seem in its ability to model cloth wrinkles and shadows, resulting in more realist renderings.

\end{document}